\documentclass[11pt]{article}
\usepackage{epsfig}
\usepackage{colacl}
\author{Mark Johnson
        \\Cognitive and Linguistic Sciences\\Brown University \And
        Stuart Geman\\Applied Mathematics\\Brown University \And
        Stephen Canon\\Cognitive and Linguistic Sciences\\Brown University \AND
        Zhiyi Chi\\Dept. of Statistics\\The University of Chicago \And
        Stefan Riezler\\Institut f\"ur Maschinelle Sprachverarbeitung\\
         Universit\"at Stuttgart}
\title{Estimators for Stochastic ``Unification-Based'' Grammars\thanks{This 
       research was
       supported by the National Science Foundation  
       (SBR-9720368), the US Army Research Office (DAAH04-96-BAA5), and
Office of Naval Research (N00014-97-1-0249).}}

\renewcommand{\Pr}{{\rm P}}
\renewcommand{\L}{{\rm L}}
\newcommand{\E}{{\rm E}}

\newcommand{\PL}{{\rm PL}}

\newcommand{\omegaseq}{{\widetilde\omega}}
\newcommand{\omegatest}{{{\widetilde\omega}_{\rm test}}}

\newcommand{\nexteqnno}{\refstepcounter{equation}\theequation}
\newcommand{\eqnno}{\theequation}

\newcommand{\Abney}{Abney~\shortcite{Abney97}}
\newcommand{\Jelinek}{Jelinek~\shortcite{Jelinek97}}
\newcommand{\Press}{Press et~al.~\shortcite{Press92}}
\newcommand{\Berger}{Berger et~al.~\shortcite{Berger96}}

\begin{document}\bibliographystyle{acl}

\maketitle

\begin{abstract}
 Log-linear models provide a statistically sound framework for
 Stochastic ``Unification-Based'' Grammars (SUBGs) and stochastic
 versions of other kinds of grammars.  We
 describe two computationally-tractable ways of estimating the
 parameters of such grammars from a training corpus of syntactic
 analyses, and apply
 these to estimate a stochastic version of Lexical-Functional Grammar. 
\end{abstract}

\section{Introduction}

Probabilistic methods have revolutionized computational linguistics.
They can provide a systematic treatment of preferences in
parsing. Given a suitable estimation procedure, stochastic models can
be ``tuned'' to reflect the properties of a corpus.  On the other
hand, ``Unification-Based'' Grammars (UBGs) can express a variety of
linguistically-important syntactic and semantic constraints.  However,
developing Stochastic ``Unification-based'' Grammars (SUBGs) has not
proved as straight-forward as might be hoped.

The simple ``relative frequency'' estimator for PCFGs yields the 
maximum likelihood parameter estimate, which is to say that it minimizes the
Kulback-Liebler divergence between the training and estimated
distributions.  On the other hand, as 
\Abney\ points out, the context-sensitive dependencies
that ``unification-based'' constraints introduce 
render the relative frequency estimator suboptimal:
in general it does not maximize the likelihood and it is inconsistent.

\Abney\ proposes a Markov Random Field or log linear model for SUBGs,
and the models described here are instances of Abney's general
framework.  However, the Monte-Carlo parameter estimation procedure
that Abney proposes seems to be computationally
impractical for reasonable-sized grammars.  Sections~\ref{s:ple}
and~\ref{s:sae} describe two new estimation procedures which are
computationally tractable.  Section~\ref{s:eval} describes an
experiment with a small LFG corpus provided to us by Xerox {\sc Parc}.
The log linear framework and the
estimation procedures are extremely general, and they apply
directly to stochastic versions of HPSG and other theories of
grammar.

\section{Features in SUBGs} \label{s:features}

\noindent
We follow the statistical literature in using the term {\em feature}
to refer to the properties that parameters are associated with (we use
the word ``attribute'' to refer to the attributes or features of a
UBG's feature structure).  Let $\Omega$ be the set of all possible
grammatical or well-formed analyses. Each feature $f$ maps a syntactic
analysis $\omega \in \Omega$ to a real value $f(\omega)$.  The form of
a syntactic analysis depends on the underlying linguistic theory.  For
example, for a PCFG $\omega$ would be parse tree, for a LFG $\omega$
would be a tuple consisting of (at least) a c-structure, an
f-structure and a mapping from c-structure nodes to f-structure
elements, and for a Chomskyian transformational grammar $\omega$ would
be a derivation.  

Log-linear models are models in which the log probability is a linear
combination of feature values (plus a constant).  
PCFGs, Gibbs distributions, Maximum-Entropy distributions and Markov
Random Fields are all examples of log-linear models.
A log-linear model associates each feature $f_j$ with a real-valued
parameter $\theta_j$.  A log-linear model with $m$ features is one in
which the likelihood $\Pr(\omega)$ of an analysis $\omega$ is:
\begin{eqnarray*}
 \Pr_\theta(\omega) & = & 
     {1 \over Z_\theta} e^{\sum_{j = 1, \ldots, m} \theta_j f_j(\omega)} 
 \\
 Z_\theta           & = & \sum_{\omega' \in \Omega} 
                  e^{\sum_{j = 1, \ldots, m} \theta_j f_j(\omega')} 
\end{eqnarray*}

While the estimators described below make no assumptions about the
range of the $f_i$, in the models considered here the value of each
feature $f_i(\omega)$ is the number of times a particular structural
arrangement or configuration occurs in the analysis $\omega$, so
$f_i(\omega)$ ranges over the natural numbers.

For example, the features of a PCFG are indexed by productions, i.e.,
the value $f_i(\omega)$ of feature $f_i$ is the number of times the
$i$th production is used in the derivation $\omega$.  This set of
features induces a tree-structured dependency graph on the productions
which is characteristic of Markov Branching Processes
\cite{Pearl88,Frey98}.  This tree structure has the important
consequence that simple ``relative-frequencies'' yield
maximum-likelihood estimates for the $\theta_i$.

Extending a PCFG model by adding additional features not associated
with productions will in general add additional dependencies,
destroy the tree structure, and substantially complicate maximum
likelihood estimation.

This is the situation for a SUBG, even if the features are production
occurences.  The unification constraints create non-local dependencies
among the productions and the dependency graph of a SUBG is usually not
a tree.  Consequently, maximum likelihood estimation is no longer a
simple matter of computing relative frequencies.  But the resulting
estimation procedures (discussed in detail, shortly), albeit more
complicated, have the virtue of applying to essentially arbitrary
features---of the production or non-production type.  That is, since
estimators capable of finding maximum-likelihood parameter estimates
for production features in a SUBG will also find maximum-likelihood
estimates for non-production features, there is no motivation for
restricting features to be of the production type.

Linguistically there is no particular reason for assuming that
productions are the best features to use in a stochastic language
model.  For example, the adjunct attachment ambiguity in (\ref{e:ex1})
results in alternative syntactic structures which use the same
productions the same number of times in each derivation, so a model
with only production features would necessarily assign them the same
likelihood.  Thus models that use production features alone predict
that there should not be a systematic preference for one of these
analyses over the other, contrary to standard psycholinguistic results.
\begin{enumerate}
\item[\nexteqnno.a\label{e:ex1}] \small
Bill thought Hillary $[_{\rm VP} [_{\rm VP} \mbox{ left }] \mbox{ yesterday }]$ 
\item[\eqnno.b] \small Bill $[_{\rm VP} [_{\rm VP} \mbox{ thought Hillary left }] \mbox{ yesterday }]$
\end{enumerate}

There are many different ways of choosing features for a SUBG, and
each of these choices makes an empirical claim about possible
distributions of sentences.  Specifying the features of a SUBG is as
much an empirical matter as specifying the grammar itself.  For any
given UBG there are a large (usually infinite) number of SUBGs that
can be constructed from it, differing only in the features that each
SUBG uses.

In addition to production features, the stochastic LFG models
evaluated below used the following kinds of features, guided by the
principles proposed by Hobbs and Bear~\shortcite{Hobbs95}. Adjunct and
argument features indicate adjunct and argument attachment
respectively, and permit the model to capture a general argument
attachment preference.  In addition, there are specialized adjunct and
argument features corresponding to each grammatical function used in
LFG (e.g., SUBJ, OBJ, COMP, XCOMP, ADJUNCT, etc.).  There are features
indicating both high and low attachment (determined by the complexity
of the phrase being attached to).  Another feature indicates
non-right-branching nonterminal nodes.  There is a feature for
non-parallel coordinate structures (where parallelism is measured in
constituent structure terms). Each f-structure attribute-atomic value
pair which appears in any feature structure is also used as a feature.
We also use a number of features
identifying syntactic structures that seem particularly important in
these corpora, such as a feature identifying NPs that are dates (it
seems that date interpretations of NPs are preferred).  We would have
liked to have included features concerning specific lexical items (to
capture head-to-head dependencies), but we felt that our corpora were
so small that the associated parameters could not be accurately estimated.

\section{A pseudo-likelihood estimator for log linear models} \label{s:ple}

\noindent
Suppose $\omegaseq = \omega_1, \ldots, \omega_n$ is a training corpus
of $n$ syntactic analyses.  Letting $f_j(\omegaseq) = \sum_{i = 1,
\ldots, n} f_j(\omega_i)$, the log likelihood of the corpus $\omegaseq$
and its derivatives are:
\begin{eqnarray}
\log \L_\theta(\omegaseq) & = &
	{  \sum_{j = 1, \ldots, m} \theta_j  f_j(\omegaseq)  } - n \log Z_\theta 
	\label{e:logL} 
 \\
\hspace*{-1em}
{\partial \log \L_\theta(\omegaseq) \over \partial \theta_j} & = &
	f_j(\omegaseq) - n \E_{\theta}(f_j)  \label{e:partial}
\end{eqnarray}
where $\E_{\theta}(f_j)$ is the expected value of $f_j$ under the
distribution determined by the parameters $\theta$.  The
maximum-likelihood estimates are the $\theta$ which maximize $\log
\L_\theta(\omegaseq)$.  The chief difficulty in finding the
maximum-likelihood estimates is calculating $\E_\theta(f_j)$,
which involves summing over the space of well-formed syntactic
structures $\Omega$.  There seems to be no analytic or efficient
numerical way of doing this for a realistic SUBG.

\Abney\ proposes a gradient ascent, based upon a Monte Carlo procedure
for estimating $\E_\theta(f_j)$.  The idea is to generate random
samples of feature structures from the distribution
$\Pr_{\hat{\theta}}(\omega)$, where $\hat{\theta}$ is the current
parameter estimate, and to use these to estimate
$\E_{\hat{\theta}}(f_j)$, and hence the gradient of the likelihood.
Samples are generated as follows: Given a SUBG, Abney constructs a
covering PCFG based upon the SUBG and $\hat{\theta}$, the current
estimate of $\theta$.  The derivation trees of the PCFG can be mapped
onto a set containing all of the SUBG's syntactic analyses.  Monte
Carlo samples from the PCFG are comparatively easy to generate, and
sample syntactic analyses that do not map to well-formed SUBG
syntactic structures are then simply discarded.  This generates a
stream of syntactic structures, but not distributed according to
$\Pr_{\hat{\theta}}(\omega)$ (distributed instead according to the
restriction of the PCFG to the SUBG).  Abney proposes using a
Metropolis acceptance-rejection method to adjust the distribution of
this stream of feature structures to achieve detailed balance, which
then produces a stream of feature structures distributed according to
$\Pr_{\hat{\theta}}(\omega)$.

While this scheme is theoretically sound, it would appear to be
computationally impractical for realistic SUBGs.  Every step of the
proposed procedure (corresponding to a single step of gradient ascent)
requires a very large number of PCFG samples: samples must be found
that correspond to well-formed SUBGs; many such samples are required
to bring the Metropolis algorithm to (near) equilibrium; many samples
are needed at equilibrium to properly estimate
$\E_{\hat{\theta}}(f_j)$.

The idea of a gradient ascent of the likelihood (\ref{e:logL}) is
appealing---a simple calculation reveals that the likelihood is
concave and therefore free of local maxima.  But the gradient (in
particular, $\E_\theta(f_j)$) is intractable.  This motivates an
alternative strategy involving a data-based estimate of
$\E_\theta(f_j)$:
\begin{eqnarray}
\E_\theta (f_j) & = & \E_\theta (\E_\theta (f_j(\omega)|y(\omega))) \\
& \approx  & \frac{1}{n}\sum_{i = 1, \ldots, n} \E_\theta(f_j(\omega)|y(\omega)=y_i)
\label{e:appox_exp}
\end{eqnarray}
where $y(\omega)$ is the yield belonging to the syntactic analysis 
$\omega$, and $y_i=y(\omega_i)$ is the yield belonging to the $i$'th
sample in the training corpus.

The point is that $\E_\theta(f_j(\omega)|y(\omega)=y_i)$ is generally
computable.  In fact, if $\Omega(y)$ is the set of well-formed
syntactic structures that have yield $y$ (i.e., the set of possible
parses of the string $y$), then
\begin{eqnarray*}
\lefteqn{\E_\theta(f_j(\omega)|y(\omega)=y_i) \; = } \\ & & 
\frac
{
\sum_{\omega' \in \Omega(y_i)}
f_j(\omega') e ^ {\sum_{k = 1, \ldots, m} \theta_k f_k(\omega')}
}
{
\sum_{\omega' \in \Omega(y_i)}
e ^ {\sum_{k = 1, \ldots, m} \theta_k f_k(\omega')}
}
\end{eqnarray*}
Hence the calculation of the conditional expectations 
only involves summing over the possible syntactic analyses or parses
$\Omega(y_i)$ of the strings in the training corpus.  While it is
possible to construct UBGs for which the number of possible parses is
unmanageably high, for many grammars it is quite manageable to enumerate
the set of possible parses and thereby directly evaluate 
$\E_\theta(f_j(\omega)|y(\omega)=y_i)$.

Therefore, we propose replacing the gradient, (\ref{e:partial}), by
\begin{equation}
f_j(\omegaseq) - \sum_{i = 1, \ldots, n}\E_\theta(f_j(\omega)|y(\omega)=y_i)
\label{e:mod_partial}
\end{equation}
and performing a gradient ascent.  Of course (\ref{e:mod_partial}) is no
longer the gradient of the likelihood function, but fortunately it is (exactly)
the gradient of (the log of) another criterion:
\begin{equation}
\PL_\theta(\omegaseq) = \prod_{i = 1, \ldots, n}\Pr_\theta(\omega=\omega_i|y(\omega)=y_i)
\label{e:pseudol}
\end{equation}
Instead of maximizing the likelihood of the syntactic analyses over the training 
corpus, we maximize the {\em conditional} likelihood of these analyses
{\em given the observed yields}.  In our experiments, we have used
a conjugate-gradient optimization program adapted from the one
presented in \Press.

Regardless of the pragmatic (computational) motivation, one could
perhaps argue that the conditional probabilities
$\Pr_\theta(\omega|y)$ are as useful (if not more useful) as the full
probabilities $\Pr_\theta(\omega)$, at least in those cases for which
the ultimate goal is syntactic analysis.  \Berger\ and \Jelinek\ make
this same point and arrive at the same estimator, albeit through a
maximum entropy argument.

The problem of estimating parameters for log-linear models is not new.
It is especially difficult in cases, such as ours, where a large
sample space makes the direct computation of expectations infeasible.
Many applications in spatial statistics, involving Markov random
fields (MRF), are of this nature as well.  In his seminal development
of the MRF approach to spatial statistics, Besag introduced a
``pseudo-likelihood'' estimator to address these difficulties
\cite{Besag74,Besag75}, and in fact our proposal here is an instance
of his method.  In general, the likelihood function is replaced by a
more manageable product of conditional likelihoods (a {\em
pseudo-likelihood}---hence the designation $\PL_\theta$), which
is then optimized over the parameter vector, instead of the likelihood
itself.  In many cases, as in our case here, this substitution side
steps much of the computational burden {\em without sacrificing
consistency} (more on this shortly).

What are the asymptotics of optimizing a pseudo-likelihood function?
Look first at the likelihood itself.  For large n:
\begin{eqnarray}
\frac{1}{n}\log \L_\theta(\omegaseq) & = & \frac{1}{n} \log \prod_{i = 1, \ldots, n}\Pr_\theta(\omega_i) \nonumber \\
 & = & \frac{1}{n} \sum_{i = 1, \ldots, n} \log \Pr_\theta(\omega_i) \nonumber \\
& \approx & \int \Pr_{\theta_o}(\omega)\log \Pr_\theta(\omega)d\omega
\label{e:kldiv}
\end{eqnarray}
where $\theta_o$ is the true (and unknown) parameter vector.  
Up to a constant, (\ref{e:kldiv}) is the negative of the 
Kullback-Leibler divergence between the true and 
estimated distributions of syntactic analyses.
As sample size grows, maximizing likelihood amounts to minimizing divergence.
As for pseudo-likelihood:
\begin{eqnarray}
\frac{1}{n}\log \PL_\theta(\omegaseq) & = & 
\frac{1}{n} \log \prod_{i = 1, \ldots, n} \Pr_\theta(\omega=\omega_i|y(\omega)=y_i) \nonumber \\
& = & \frac{1}{n} \sum_{i = 1, \ldots, n} \log \Pr_\theta(\omega=\omega_i|y(\omega)=y_i) \nonumber \\
& \approx & \E_{\theta_o}[\int \Pr_{\theta_o}(\omega|y)\log \Pr_\theta(\omega|y)d\omega] \nonumber
\end{eqnarray}
So that maximizing pseudo-likelihood (at large samples) amounts to
minimizing the average (over yields) divergence between the true and
estimated {\em conditional distributions of analyses given yields}.

Maximum likelihood estimation is consistent: under broad conditions
the sequence of distributions $\Pr_{{\hat{\theta}}_n}$, associated
with the maximum likelihood estimator for $\theta_o$ given the samples
$\omega_1,\ldots\omega_n$, converges to $\Pr_{\theta_o}$.
Pseudo-likelihood is also consistent, but in the present
implementation it is consistent for the conditional distributions
$\Pr_{\theta_o}(\omega|y(\omega))$ and not necessarily for the full
distribution $\Pr_{\theta_o}$ (see Chi\ \shortcite{Chi98}).  It is not
hard to see that pseudo-likelihood will not always correctly estimate
$\Pr_{\theta_o}$. Suppose there is a feature $f_i$ which depends only
on yields: $f_i(\omega)=f_i(y(\omega))$.  (Later we will refer to such
features as {\em pseudo-constant}.)  In this case, the derivative of
$\PL_\theta(\omegaseq)$ with respect to $\theta_i$ is zero; ${\rm
PL}_\theta(\omegaseq)$ contains no information about $\theta_i$.  In
fact, in this case {\em any} value of $\theta_i$ gives the {\em same}
conditional distribution $\Pr_\theta(\omega|y(\omega))$; $\theta_i$ is
irrelevant to the problem of choosing good parses.

Despite the assurance of consistency, pseudo-likelihood estimation is
prone to over fitting when a large number of features is matched
against a modest-sized training corpus.  One particularly troublesome
manifestation of over fitting results from the existence of features
which, relative to the training set, we might term ``pseudo-maximal'':
Let us say that a feature $f$ is {\em pseudo-maximal} for a yield $y$
iff $\forall \omega' \in \Omega(y) \, f(\omega) \geq f(\omega')$ where
$\omega$ is any correct parse of $y$, i.e., the feature's value on
every correct parse $\omega$ of $y$ is greater than or equal to its
value on any other parse of $y$.  Pseudo-minimal features are defined
similarly.  It is easy to see that if $f_j$ is pseudo-maximal on {\em
each sentence} of the training corpus then the parameter assignment
$\theta_j = \infty$ maximizes the corpus pseudo-likelihood.
(Similarly, the assignment $\theta_j = -\infty$ maximizes
pseudo-likelihood if $f_j$ is pseudo-minimal over the training
corpus).  Such infinite parameter values indicate that the model
treats pseudo-maximal features categorically; i.e., any parse with a
non-maximal feature value is assigned a zero conditional probability.

Of course, a feature which is pseudo-maximal over the training corpus
is not necessarily pseudo-maximal for all yields.  This is an instance
of over fitting, and it can be addressed, as is customary, by adding a
regularization term that promotes small values of $\theta$ to the
objective function.  A common choice is to add a quadratic to the
log-likelihood, which corresponds to multiplying the likelihood itself
by a normal distribution.  In our experiments, we multiplied the
pseudo-likelihood by a zero-mean normal in $\theta_1,\ldots\theta_m$,
with diagonal covariance, and with standard deviation $\sigma_j$ for
$\theta_j$ equal to $7$ times the maximum value of $f_j$ found in any
parse in the training corpus.  (We experimented with other values for
$\sigma_j$, but the choice seems to have little effect).  Thus instead
of maximizing the log pseudo-likelihood, we choose $\hat{\theta}$ to
maximize
\begin{equation} \label{e:logPLbias}
\log \PL_\theta(\omegaseq) - \sum_{j = 1, \ldots, m} { \theta_j ^ 2 \over 2 \sigma_j ^ 2}
\end{equation}

\section{A maximum correct estimator for log linear models} \label{s:sae}

\noindent
The pseudo-likelihood estimator described in the last section finds
parameter values which maximize the conditional probabilities of the
observed parses (syntactic analyses) given the observed sentences
(yields) in the training corpus.  One of the empirical evaluation
measures we use in the next section measures the number of correct
parses selected from the set of all possible parses.  This suggests
another possible objective function: choose $\hat{\theta}$ to maximize
the number $C_\theta(\omegaseq)$ of times the maximum likelihood parse
(under $\theta$) is in fact the correct parse, in the training corpus.

$C_\theta(\omegaseq)$ is a highly discontinuous function of $\theta$,
and most conventional optimization algorithms perform poorly on it.
We had the most success with a slightly modified version of the
simulated annealing optimizer described in \Press.  This procedure is
much more computationally intensive than the gradient-based pseudo-likelihood
procedure.  Its computational difficulty grows (and the quality of
solutions degrade) rapidly with the number of features.

\section{Empirical evaluation} \label{s:eval}

\noindent
Ron Kaplan and Hadar Shemtov at Xerox {\sc Parc} provided us with two
LFG parsed corpora.  The Verbmobil corpus contains appointment
planning dialogs, while the Homecentre corpus is drawn from Xerox
printer documentation.  Table~\ref{t:corpora} summarizes the basic
properties of these corpora.  These corpora contain packed
c/f-structure representations \cite{Maxwell95b} of the grammatical
parses of each sentence with respect to Lexical-Functional grammars.
The corpora also indicate which of these parses is in fact the correct
parse (this information was manually entered).  Because 
slightly different grammars were used for each corpus we chose not to
combine the two corpora, although we used the set of features
described in section~\ref{s:features} for both in the experiments
described below.  Table~\ref{t:features} describes the properties
of the features used for each corpus.

\begin{table*}
\begin{center}
\begin{tabular}{lcc}
	& \bf Verbmobil corpus & \bf Homecentre corpus \\
Number of sentences & 540 & 980 \\
Number of ambiguous sentences & 314 & 481 \\
Number of parses of ambiguous sentences & 3245 & 3169 
\end{tabular}
\end{center}
\caption{Properties of the two corpora used to evaluate the estimators.} \label{t:corpora}
\end{table*}

\begin{table*}
\begin{center}
\begin{tabular}{lcc}
	& \bf Verbmobil corpus & \bf Homecentre corpus \\
Number of features & 191 & 227 \\
Number of rule features & 59 & 57 \\
Number of pseudo-constant features & 19 & 41 \\
Number of pseudo-maximal features & 12 & 4 \\
Number of pseudo-minimal features & 8  & 5
\end{tabular}
\end{center}
\caption{Properties of the features used in the stochastic LFG models. The numbers
of pseudo-maximal and pseudo-minimal features do not include pseudo-constant features.} \label{t:features}
\end{table*}

In addition to the two estimators described above we also present
results from a baseline estimator in which all parses are treated
as equally likely (this corresponds to setting all the
parameters $\theta_j$ to zero).  

We evaluated our estimators using held-out test corpus $\omegatest$.
We used two evaluation measures.  In an actual parsing application a
SUBG might be used to identify the correct parse from the set of
grammatical parses, so our first evaluation measure counts the number
$C_{\hat{\theta}}(\omegatest)$ of sentences in the test corpus
$\omegatest$ whose maximum likelihood parse under the estimated model
$\hat{\theta}$ is actually the correct parse.  If a sentence has $l$
most likely parses (i.e., all $l$ parses have the same
conditional probability) and one of these parses is the correct parse, then
we score $1/l$ for this sentence.

The second evaluation measure is the pseudo-likelihood itself,
$\PL_{\hat{\theta}}(\omegatest)$.  The
pseudo-likelihood of the test corpus is the likelihood of the correct
parses given their yields, so pseudo-likelihood measures how much of
the probability mass the model puts onto the correct analyses.  This
metric seems more relevant to applications where the system needs to
estimate how likely it is that the correct analysis lies in a certain
set of possible parses; e.g., ambiguity-preserving translation and
human-assisted disambiguation.
To make the numbers more manageable, we actually present the negative
logarithm of the pseudo-likelihood rather than the pseudo-likelihood
itself---so smaller is better.

\begin{table*}
\begin{center}
\begin{tabular}{lcccc}
  & \multicolumn{2}{c}{\bf Verbmobil corpus} & \multicolumn{2}{c}{\bf Homecentre corpus} \\
  & \multicolumn{1}{c}{$C(\omegatest)$} & \multicolumn{1}{c}{$- \log \PL(\omegatest) $} &
    \multicolumn{1}{c}{$C(\omegatest)$} & \multicolumn{1}{c}{$- \log \PL(\omegatest) $} \\
Baseline estimator & 9.7\% & 533 & 15.2\% & 655 \\
Pseudo-likelihood estimator & 58.7\% & 396 & 58.8\% & 583 \\
Correct-parses estimator & 53.7\% & 469 & 53.2\% & 604 
\end{tabular}
\end{center}
\caption{An empirical evaluation of the estimators.  {$C(\omegatest)$}
is the number of maximum likelihood parses of the test corpus that
were the correct parses, and {$- \log \PL(\omegatest) $} is the
negative logarithm of the pseudo-likelihood of the test corpus.}
\label{t:results}
\end{table*}

Because of the small size of our corpora we evaluated our estimators
using a 10-way cross-validation paradigm.  We randomly assigned
sentences of each corpus into 10 approximately equal-sized subcorpora,
each of which was used in turn as the test corpus.  We evaluated on
each subcorpus the parameters that were estimated from the 9 remaining
subcorpora that served as the training corpus for this run.  The
evaluation scores from each subcorpus were summed in order to provide
the scores presented here.

Table~\ref{t:results} presents the results of the empirical
evaluation.  The superior performance of both estimators on the
Verbmobil corpus probably reflects the fact that the non-rule features
were designed to match both the grammar and content of that corpus.
The pseudo-likelihood estimator performed better than the
correct-parses estimator on both corpora under both evaluation
metrics.  There seems to be substantial over learning in all these
models; we routinely improved performance by discarding features.
With a small number of features the correct-parses estimator typically
scores better than the pseudo-likelihood estimator on the
correct-parses evaluation metric, but the pseudo-likelihood estimator
always scores better on the pseudo-likelihood evaluation metric.

\section{Conclusion}

\noindent
This paper described a log-linear model for SUBGs and evaluated two
estimators for such models.  Because estimators that can estimate rule
features for SUBGs can also estimate other kinds of features, there is
no particular reason to limit attention to rule features in a SUBG.
Indeed, the number and choice of features strongly influences the
performance of the model.  The estimated models are able to identify
the correct parse from the set of all possible parses approximately
$50\%$ of the time.

We would have liked to introduce features corresponding to
dependencies between lexical items.  Log-linear models are well-suited
for lexical dependencies, but because of the large number of such
dependencies substantially larger corpora will probably be needed to
estimate such models.\footnote{Alternatively, it may be possible to
use a simpler non-SUBG model of lexical dependencies estimated from a
much larger corpus as the reference distribution with respect to which
the SUBG model is defined, as described in \Jelinek.}

However, there may be applications which can benefit from a model that
performs even at this level.  For example, in a machine-assisted
translation system a model like ours could be used to order possible
translations so that more likely alternatives are presented before
less likely ones.  In the ambiguity-preserving translation framework,
a model like this one could be used to choose between sets of analyses
whose ambiguities cannot be preserved in translation.

\bibliography{mj}

\begin{thebibliography}{}

\bibitem[\protect\citename{Abney}1997]{Abney97}
Steven~P. Abney.
\newblock 1997.
\newblock {S}tochastic {A}ttribute-{V}alue {G}rammars.
\newblock {\em Computational Linguistics}, 23(4):597--617.

\bibitem[\protect\citename{Berger \bgroup et al.\egroup }1996]{Berger96}
Adam~L. Berger, Vincent~J. {Della Pietra}, and Stephen~A. {Della Pietra}.
\newblock 1996.
\newblock A maximum entropy approach to natural language processing.
\newblock {\em Computational Linguistics}, 22(1):39--71.

\bibitem[\protect\citename{Besag}1974]{Besag74}
J.~Besag.
\newblock 1974.
\newblock Spatial interaction and the statistical analysis of lattice systems
  (with discussion).
\newblock {\em Journal of the Royal Statistical Society, Series D},
  36:192--236.

\bibitem[\protect\citename{Besag}1975]{Besag75}
J.~Besag.
\newblock 1975.
\newblock Statistical analysis of non-lattice data.
\newblock {\em The Statistician}, 24:179--195.

\bibitem[\protect\citename{Chi}1998]{Chi98}
Zhiyi Chi.
\newblock 1998.
\newblock {\em Probability Models for Complex Systems}.
\newblock {Ph.D.} thesis, Brown University.

\bibitem[\protect\citename{Frey}1998]{Frey98}
Brendan~J. Frey.
\newblock 1998.
\newblock {\em Graphical Models for Machine Learning and Digital
  Communication}.
\newblock The MIT Press, Cambridge, Massachusetts.

\bibitem[\protect\citename{Hobbs and Bear}1995]{Hobbs95}
Jerry~R. Hobbs and John Bear.
\newblock 1995.
\newblock Two principles of parse preference.
\newblock In Antonio Zampolli, Nicoletta Calzolari, and Martha Palmer, editors,
  {\em Linguistica Computazionale: Current Issues in Computational Linguistics:
  In Honour of Don Walker}, pages 503--512. Kluwer.

\bibitem[\protect\citename{Jelinek}1997]{Jelinek97}
Frederick Jelinek.
\newblock 1997.
\newblock {\em Statistical Methods for Speech Recognition}.
\newblock The MIT Press, Cambridge, Massachusetts.

\bibitem[\protect\citename{Maxwell~III and Kaplan}1995]{Maxwell95b}
John~T. Maxwell~III and Ronald~M. Kaplan.
\newblock 1995.
\newblock A method for disjunctive constraint satisfaction.
\newblock In Mary Dalrymple, Ronald~M. Kaplan, John~T. Maxwell~III, and Annie
  Zaenen, editors, {\em Formal Issues in {L}exical-{F}unctional Grammar},
  number~47 in CSLI Lecture Notes Series, chapter~14, pages 381--481. CSLI
  Publications.

\bibitem[\protect\citename{Pearl}1988]{Pearl88}
Judea Pearl.
\newblock 1988.
\newblock {\em Probabalistic Reasoning in Intelligent Systems: Networks of
  Plausible Inference}.
\newblock Morgan Kaufmann, San Mateo, California.

\bibitem[\protect\citename{Press \bgroup et al.\egroup }1992]{Press92}
William~H. Press, Saul~A. Teukolsky, William~T. Vetterling, and Brian~P.
  Flannery.
\newblock 1992.
\newblock {\em Numerical Recipies in C: The Art of Scientific Computing}.
\newblock Cambridge University Press, Cambridge, England, 2nd edition.

\end{thebibliography}
\end{document}